\documentclass[11pt]{article}
\usepackage{geometry}
\usepackage{coling2020}
\usepackage{times}
\usepackage{url}
\usepackage{latexsym}
\usepackage{microtype}

\usepackage{xcolor}
\usepackage{tabto}
\usepackage{pdfpages}

\usepackage{array}
\newcolumntype{C}[1]{>{\centering\arraybackslash}p{#1}}

\colingfinalcopy 

\title{SemEval-2020 Task 5: Counterfactual Recognition}

\author{Xiaoyu Yang$^\dagger$ \qquad
Stephen Obadinma$^\dagger$  \qquad 
Huasha Zhao$^\S$ \qquad \\
\textbf{Qiong Zhang}$^\S$ \qquad 
\textbf{Stan Matwin}$^\ddagger$ \qquad
\textbf{Xiaodan Zhu}$^\dagger$ \\ 
  $^\dagger$ Ingenuity Labs Research Institute \& ECE, Queen's University, Canada \\ 
  $^\S$ Alibaba Group, San Mateo, CA \\ 
  $^\ddagger$ Department of Computer Science, Dalhousie University, Canada \\
  \texttt{\{xiaoyu.yang, 16sco, xiaodan.zhu\}@queensu.ca} \\
  \texttt{qz.zhang@alibaba-inc.com} \quad
  \texttt{creislerzhao@gmail.com}\\
  \texttt{stan@cs.dal.ca} \quad
  }

\date{}

\begin{document}
\maketitle
\begin{abstract}
We present a counterfactual recognition (CR) task, the shared Task 5 of SemEval-2020. 
Counterfactuals describe potential outcomes (consequents) produced by actions or circumstances that did not happen or cannot happen and are counter to the facts (antecedent). 
Counterfactual thinking is an important characteristic of the human cognitive system; it connects \textit{antecedents} and \textit{consequents} with causal relations.
Our task provides a benchmark for counterfactual recognition in natural language with two subtasks. Subtask-1 aims to determine whether a given sentence is a counterfactual statement or not. Subtask-2 requires the participating systems to extract the antecedent and consequent in a given counterfactual statement.
During the SemEval-2020 official evaluation period, we received 27 submissions to Subtask-1 and 11 to Subtask-2.
The data, baseline code, and leaderboard can be found at~\url{https://competitions.codalab.org/competitions/21691}. The data and baseline code are also available at~\url{https://zenodo.org/record/3932442}. 
\end{abstract}


\section{Introduction}
\label{intro}

Counterfactual statements describe events that did not happen or cannot happen, and the possible consequences had those events happened,
e.g., ``\textit{if kangaroos had no tails, they would topple over}''~\cite{lewis2013counterfactuals}. 
By developing a connection between the antecedent (e.g., ``\textit{kangaroos had no tails}'') and consequent (e.g., ``\textit{they would topple over}''), based on the imagination of possible worlds, humans can naturally form some causal judgments; e.g., having tails can prevent kangaroos from toppling over. 
One can understand counterfactuals using knowledge and explore the relationship between causes and effects. Although we may not be able to rollback the events which have happened or make impossible events occur in the real world, we can still think of potential outcomes of alternatives.

Counterfactual thinking is a remarkable ability of human beings and is considered by many researchers, to act as the highest level of causation in the ladder of causal reasoning. 
Even the most advanced artificial intelligence system may still be far from achieving human-like counterfactual reasoning. 
Counterfactual reasoning is an important component for AI systems in obtaining stronger capability in generalization~\cite{pearl2018book}.

Modeling counterfactuals has been studied in many different disciplines. 
For example, research in psychology has shown that counterfactual thinking can affect human cognition and behaviors~\cite{epstude2008functional,kray2010might}. 
The landmark paper of ~\cite{goodman1947problem} gives a detailed analysis of counterfactual conditionals in philosophy and logistics. As another example, counterfactuals have also been investigated in epidemiology to reveal the relationship between certain diseases and potential risk factors for those diseases ~\cite{vandenbroucke2016causality,krieger2016tale}.

We present a counterfactual recognition (CR) task, the task of determining whether a given statement conveys counterfactual thinking or not, and further analyzing the causal relations indicated by counterfactual statements. In our counterfactual recognition task, we aim to model counterfactual semantics and reasoning in natural language. Specifically, we provide a benchmark for counterfactual recognition with two subtasks. Subtask-1 requires systems to determine whether a given statement is counterfactual or not. The counterfactual detection task can serve as a foundation for downstream counterfactual analysis.
Subtask-2 requires systems to further locate the antecedent and consequent text spans in a given counterfactual statement, as the connection between an antecedent and consequent can reveal core causal inference clues. 

To build the dataset for counterfactual recognition, we extract
over 60,000 candidate 
counterfactual statements by scanning through news reports in three domains: finance, politics, and healthcare. 
The first round of annotation focuses on labeling each sample as \textit{true} or \textit{false}, where \textit{true} denotes a sample is counterfactual and \textit{false} otherwise in Subtask-1. A portion of samples labeled as \textit{true} will be further used in Subtask-2 to detect the text spans that describe the antecedent and consequent. Specifically, we carefully select 20,000 high-quality samples from the 60,000 statements and use them in Subtask-1, with 13,000 (65\%) as the training set and the rest for testing. The dataset for Subtask-2 contains 5,501 samples, among which we use 3,551 (65\%) for training and the rest for testing. 

To achieve a decent performance in our shared task, we expect the systems should have a certain level of language understanding capacity in both semantics and syntax, together with a certain level of commonsense reasoning ability. 

In Subtask-1, the top-ranked submissions all use pre-trained neural models, which appear to be an effective way to integrate knowledge learned from large corpus.  
All of these models use neural networks, which further confirms the effectiveness of distributed representation and subsymbolic approaches for this task. 
Some top systems also successfully incorporate rules to further improve the performance, suggesting the benefits of combining neural networks with symbolic approaches. 
The first-place model also utilizes data augmentation to further improve system performance. In Subtask-2, top systems take two main approaches: sequence labelling or question answering. Same as systems in Subtask-1, all of them benefit from pre-training. We will provide a more detailed analysis in the system and result section.

We built a dataset for this shared task from scratch.
Our data, baseline code, and leaderboard can be found at~\url{https://competitions.codalab.org/competitions/21691}. The data and baseline code are also available at~\url{https://zenodo.org/record/3932442}. 
In general, our task here is a relatively basic one in counterfactual analysis in natural language. We hope it will intrigue and facilitate further research on counterfactual analysis and can benefit other related downstream tasks.

%
%
\blfootnote{
    %
    %
    %
    %
    
    
    \hspace{-0.65cm}  
    This work is licensed under a Creative Commons 
    Attribution 4.0 International License.
    License details:
    \url{http://creativecommons.org/licenses/by/4.0/}.
}

\section{Task Setup}
\label{sec: task}

In this section, we detail the two counterfactual recognition subtasks and the metrics used to evaluate the performance. During the evaluation, participants can work on both subtasks or any one of them. 

\subsection{Subtask-1: Recognizing Counterfactual Statements (RCS)}

We formulate the Subtask-1 as a binary classification problem which asks the participating systems to detect whether a particular sentence is counterfactual or not. 
Below are two examples of counterfactual statements that need to be recognized:
\begin{itemize}
    \item \textbf{Example-1:} \textit{Officials say if they had authority to shut non-bank firms, the collapse of Lehman Brothers, which touched off the most virulent phase of the credit crisis, could have been avoided.}
    \item \textbf{Example-2:} \textit{The delivery numbers would have been high had it not been for the restrictions imposed by the military for security reasons.}
\end{itemize}


\subsection{Subtask-2: Detecting Antecedent and Consequent (DAC)}

Indicating causal relationships is an inherent characteristic of counterfactuals. To further detect the causal knowledge conveyed in counterfactual statements, Subtask-2 aims to extract the antecedents and consequents.
Specifically, given a counterfactual statement, systems for Subtask-2 need to identify the indices of the characters which indicate the start and end positions for antecedent and consequent, in terms of character indices: \textit{antecedent\_start\_ind, antecedent\_end\_ind, consequent\_start\_ind, consequent\_end\_ind}.
For some statements, the consequents may not be expressed in the statements, then the corresponding \textit{consequent\_start\_ind} and \textit{consequent\_end\_ind} will be set as $-1$. 
\begin{itemize}
    \item \textbf{Example-3:} \textit{The delivery numbers would have been high had it not been for the restrictions imposed by the military for security reasons.}\\
    \textit{\textbf{Antecedent:}} \textit{had it not been for the restrictions imposed by the military for security reasons}\\
    \textit{\textbf{Consequent:}} \textit{the delivery numbers would have been high}\\
    \textit{\textbf{Label:} $42, 122, 0, 40$} 
    \item \textbf{Example-4:} \textit{It should have been feasible right after the collapse of Lehman Brothers (in 2008), when prices were cheap.}\\
    \textit{\textbf{Antecedent:} \textit{it should have been feasible right after the collapse of Lehman Brothers (in 2008)}} \\
    \textit{\textbf{Consequent:}} \textit{None} \\
    \textit{\textbf{Label: $0, 81, -1, -1$}}
\end{itemize}

A counterfactual statement can be converted to a contrapositive with a true antecedent and consequent, by assuming the antecedent and consequent in the original counterfactual statement is inalterably false~\cite{goodman1947problem}. Consider the Example-3 above. It can be transposed into \textit{``since the restrictions imposed by the military for security reasons, the delivery numbers were not high''.}
After extracting the antecedent and the corresponding consequent from a counterfactual statement, we may derive a contrapositive by performing an appropriate transformation, which can naturally reveal a causal relationship between the two parts or even further indicate the properties of each part. In this way, it is possible to extract causal knowledge across corpora.


\subsection{Evaluation Metrics}

Subtask-1 is a binary classification problem evaluated with \textbf{Precision}, \textbf{Recall}, and \textbf{F1 score}~\footnote{In the official evaluation period of Subtask-1, the ranking is based on F1.}. In Subtask-2, we utilize two metrics: (i) \textbf{Exact match} is used to evaluate the percentage of predictions that exactly match the ground truth boundaries of the antecedents and consequents. (ii) \textbf{F1 score} is used to measure the overlap between the predictions and ground truth spans. For each sample, we calculate the number of tokens in the overlapped intervals by comparing the predictions and ground truth indices of antecedent and consequent boundaries. Then we can compute precision, recall, and F1 score for each sample. We take the average F1 score across all the samples in the test set.
Note the F1 score used in both subtasks is calculated as: $F1 = \frac{2*Precision*Recall}{Precision+Recall}$.


\section{Data Development}
\label{sec: data}
We develop our dataset from news articles in the finance, politics, or healthcare domain. The data development consists of data collection and annotation. We use different approaches to ensure the quality of the data.

\subsection{Data Collection}
There are two major challenges in our data construction process. 
First, due to the relative sparsity of counterfactual statements in the text, manually annotating each sentence in the original text is not of time and financial efficiency. Accordingly, we perform a filtering step to narrow down candidates.
The second challenge is rooted in the flexibility and complexity of counterfactual expressions. Not all counterfactual statements follow certain patterns, e.g., the \textit{``if + past perfect''} pattern (although this is a good pattern which can indicate a conditional relationship between the antecedent and the potential consequent). To solve these problems, we create a set of templates considering the trade-off between the effectiveness of filtering and its diversity in finding candidate counterfactuals, without making the filtering stage too rigorous.

\paragraph{Token-based Filtering} The template set consists of two subsets that jointly work to find candidate potential counterfactual statements when they are used to search through news articles. The first subset focuses on word token patterns and the second subset leverages POS tag-based patterns. 
The full list of token-based patterns are listed in Appendix A. 
Some of the patterns are based on the previous research which revealed common counterfactual constructions~\cite{hobbs2005toward,son2017recognizing,rouvoli2019if}.

\paragraph{POS-based Filtering} The second subset of templates utilize patterns based on part-of-speech tags. We identified five counterfactual forms based on~\cite{janockocounterfactuals} and coverted them into POS-based patterns to increase the chances of identifying true counterfactual statements. The details of the POS-based rules are presented in Appendix B. 
To apply the rules, we tokenize each sentence and conduct POS tagging with the NLTK library~\cite{bird2009natural}.
Then we extract the sentences which match one of the pre-defined patterns.

By applying both the token-based and POS-based rules, we obtain the candidate statements for further human annotation.

\subsection{Annotation}
As described above, each sample in Subtask-1 is labeled  either as \textit{true} (counterfactual) or \textit{false} (non-counterfactual). We employ a two-step annotation strategy. First, each sample in the candidate statement set is annotated by five annotators to determine whether it is a counterfactual statement or not. We include those annotated as \textit{true} (counterfactuals) by all five annotators, i.e., with an agreement rate of 100\%.
For negative samples (non-counterfactual statements), we take all of those labeled as \textit{false} with 100\% agreement and some sentences with 80\% agreement, which 4 out of the 5 annotators label as \textit{false}.

We use the Amazon Mechanical Turk (AMT) platform for our annotation, by splitting the samples into HITs (Human Intelligent Task, where each HIT contains 20 to 30 samples) and distributing these HITs to qualified annotators along with thorough instructions and examples.

In subtask 2, a portion of counterfactual statements (labeled as \textit{true} in Subtask-1) are further annotated, in which the text spans of antecedents and consequents in counterfactual statements are obtained. In this stage, each sample was annotated by a single annotator on Amazon Mechanical Turk, and the annotators were asked to double-check whether the sample is a counterfactual statement before underlining specific spans. 
All the samples are further manually checked by ourselves to ensure the antecedent and consequent spans are appropriately labelled by a consistent standard.


\subsection{Quality Control}
We further make the following efforts to control the quality of our datasets. 
First we set additional requirements when inviting workers to perform annotation. We only invite workers from English-speaking countries and only if the approval rates of their previous HITs are above 82\%. In addition, workers take a qualification test before starting their annotation work. The test provides detailed instructions and examples and includes 40 samples for workers to label which of the samples are counterfactuals. Using this method, we have over 70 qualified workers for our data annotation task. Having a stable pool of trained workers is beneficial for ensuring the quality of annotation. 
In the entire process of annotation, we randomly select some HITs to evaluate the accuracy and the performance of workers to justify whether to accept them or not. 

For subtask 2, we manually check all of the samples to ensure: (i) the samples are counterfactual statements (any incorrectly labelled statements are further removed from both Subtask-1 and Subtask-2 datasets); (ii) for a very small number of statements, if the antecedent and consequent spans labelled by the Turkers are not full constituent phrases, we manually adjust the span to make them full phrases. 


\subsection{Data Statistics}
Table~\ref{table: statistics task-1} shows the statistics of the data used in Subtask-1. We obtain 20,000 statements in total and we randomly split them into the training (65\%) and test set (35\%). The participants can select their own development set from the training set or use cross validation to develop their models. The training set includes 1,454 counterfactual statements and the test set includes 738 true counterfactual statements. 
The number of statements is balanced among the three domains---each domain has roughly one third of candidate statements. 
Table~\ref{table: statistics task-2} shows the size of data used in Subtask-2. In total, we have 5,501 samples and randomly split them into the training and test set. Specifically, 3,551 samples are for training and the rest for testing. Not all counterfactuals have both an antecedent and a consequent, so we also provide statistics for samples that have only antecedents, and those that have both antecedents and consequents. Figure~\ref{fig:statistics} in Appendix C shows statistics for the frequency of the number of words in both the Subtask-1 and Subtask-2 training and test datasets. 


\begin{table*}[t]
\centering
\begin{tabular}{l|ccc}
\hline
Dataset      & Counterfact. & Non-counterfact. & Total    \\ \hline
Train  & 1,454            & 11,546 & 13,000               \\
Test       & \phantom{0,}738              & \phantom{0}6,262 & \phantom{0}7,000    
\\ \hline
Total  & 2,192 & 17,808 & 20,000
\\ \hline
\end{tabular}
\caption{Sizes of the training and test set used in Subtask-1.}
\label{table: statistics task-1}
\end{table*}

\begin{table*}[t]
\centering
\begin{tabular}{l|ccc}
\hline
Dataset      & Antecedent only & Ante. \& Cons. & Total   \\ \hline
Train & 520             & 3,031 & 3,551                    \\
Test     & 268             & 1,682 & 1,950                    \\ \hline
Total      & 788             & 4,713 & 5,501                     \\ \hline
\end{tabular}
\caption{Sizes of the training and test set used in Subtask-2. \textit{Antecedent only} are statements that only have antecedents but not consequents, and \textit{Ante. \& Cons.} means statements having both antecedents and consequents.}
\label{table: statistics task-2}
\end{table*}



\section{Systems and Results}
\label{sec: submission-1}


\subsection{Subtask 1: Recognizing Counterfactual Statements (RCS)}

The baseline used for Subtask-1 is a simple SVM classifier with the linear kernel function. 
In the baseline model, we first take some basic preprocessing steps starting with lemmatization; we then extract term frequency and inverse document frequency (tf-idf) features for training the SVM model to perform binary classification. The motivation behind using this simple SVM as a baseline is to create a simple model that can identify counterfactuals by learning and searching for the presence keywords and phrases like ``\textit{had}'' or ``\textit{should have been}'', that tend to mark the presence of a counterfactual in a number of counterfactual grammatical forms. The baseline has poor performance, signalling that most counterfactuals cannot be determined based on the presence of certain words and that reasoning is necessary.
The baseline is not shown on the official leaderboard but in Table~\ref{table: results task-1}.


We received 27 submissions to Subtask-1. Table~\ref{table: results task-1} shows all the official submission results and nearly all of them exceed the performance of the provided baseline model.
The top-ranked submissions all use pre-trained neural models, which have achieved the state-of-the-art results across many natural language processing (NLP) tasks~\cite{devlin2018bert,radford2018improving,radford2019language,yang2019xlnet,liu2019roberta,lan2019albert}, which we believe is an effective way to integrate additional external knowledge that does not exist in the training data, including common sense. 
Some participants like \textit{shngt} experimented with classic machine learning methods like SVM and gradient boosted random forests, and found that model performance plateaued at an F1 score of around 60 percent ~\cite{shngt2020SemevalTask5}, showing that these methods cannot capture counterfactual reasoning as well as the pre-trained models. One team, \textit{Serena}, use a non-transformer approach, basing their system on Ordered Neurons LSTM (ON-LSTM) with Hierarchical Attention Network (HAN) and a Pooling operation is done for dimensionality reduction ~\cite{Serena2020SemevalTask5}. Their system struggles with data imbalance and they conclude that transformer networks can improve their performance. Among the top 8 competitors, BERT and RoBERTa based systems are most popular, being used as the primary models or as a part of their final ensemble in  5 and 4 of the top 8 participants' systems respectively. XLNet and ALBERT are less popular choices, but they are also used in the first and second of the top 8 participating systems respectively.

In addition, the top models adopt ensemble strategies, and in most cases, achieve better performance than that of individual classifiers.
One of the teams, \textit{shgnt}, also found using a convolutional neural network model with GloVe embeddings in their ensemble helped to enhance it~\cite{shngt2020SemevalTask5}, showing that non-transformer network based methods can be useful. Despite the commonality of using pre-trained models, many of the participating systems differ in the structures they add above the pre-trained models, to capture additional information. Rather than adding a fully connected layer on top, some competitors reconstruct the top structure of the pre-trained models or add a neural network on top. For example, to capture local patterns in counterfactual statements, \textit{Roger} ~\cite{Roger2020SemevalTask5} and \textit{shngt} ~\cite{shngt2020SemevalTask5} add a CNN before classification in some of their systems. Similarly, \textit{Baiyang2581} experiments with this upper structure and use a bidirectional GRU and bidirectional LSTM in some of their systems after the transformer network ~\cite{baiyang25812020SemevalTask5}. In contrast to modifying the upper structure of transformer networks, the fifth-place team, \textit{lenyabloko}, uses rule-based specialist modules by combining fine-tuned pre-trained models with constituency and dependency parsers to compensate for deficiencies in deep learning methods for causal inference in language to great effect ~\cite{lenyabloko2020SemevalTask5}.

The dataset is highly imbalanced in favour of non-counterfactuals, and many participants use techniques to deal with this imbalance. A range of techniques like pseudo-labelling used by \textit{haodingkui} ~\cite{haodingkui2020SemevalTask5}, multi sample dropout used by Ferryman ~\cite{Ferryman2020SemevalTask5}, and oversampling and undersampling used by some other teams, notably \textit{ad6398} who found that undersampling non-counterfactuals optimized the performance of their models ~\cite{ad63982020SemevalTask5}. \textit{changshivek} experimented with 2 novel forms of data augmentation to increase the number of counterfactual samples ~\cite{changshivek2020SemevalTask5}. The first was back translation, which involves taking counterfactual samples and translating them into another language and then translating them back in English and adding them to the dataset. The second technique was Easy Data Augmentation (EDA), namely synonym replacement of words in the counterfactual samples while making sure to preserve words relating to counterfactuals like ``should''. Back translation yielded poor performance, but decent improvement was seen when using EDA showing that this could be a viable method. These methods are all done to combat overfitting. K-fold cross validation is also a common strategy utilized by many of the participating systems to deal with the relatively small, highly imbalanced dataset to reduce some of the bias. 

Lastly, many teams experimented with pre-processing the data. \textit{Baiyang251} notes that minimal pre-processing (e.g. deleting punctuation, making sentences all lower-case) on the data yields the best results as they theorize removing these results in the loss of information useful for prediction ~\cite{baiyang25812020SemevalTask5}. Other groups note that extensive pre-processing does not yield notable performance improvements either and can even slightly hurt performance.

The best F1 score in Subtask-1, 90.9\%, was achieved by \textit{haodingkui} ~\cite{haodingkui2020SemevalTask5}. In their approach, a pseudo-labelling strategy is used to generate more data to alleviate overfitting during training. In this strategy, if all classifiers agree on the labels of certain samples in the test set, then those samples will also be used for training. For model ensembling, they incorporate BERT, RoBERTa, and XLNet. The second-place system from~\textit{josefjon} concludes that using an ensemble of RoBERTa large models performs better than any other pre-trained model ~\cite{josefjon2020SemevalTask5}. The third-place system, proposed by~\textit{Roger}, incorporates convolutional neural networks to capture strong local context information in addition to fine-tuning pre-trained models. Furthermore, it theorizes about the effectiveness of using knowledge-enriched transformers to improve performance on the task ~\cite{Roger2020SemevalTask5}. Some top systems also successfully incorporate rules to further improve the performance, suggesting the benefits of combining neural nets with symbolic approaches. 

\paragraph{Further Analysis and Challenges} In general, one of the main challenges of Subtask-1 is that identifying counterfactuals requires inference and reasoning based on common sense and knowledge. Particularly, the fact that counterfactuals often do not follow specific grammatical rules makes such an ability important for some statements. The imbalanced nature of the dataset in Subtask-1 is another challenge; therefore different methods have been proposed to address this issue such as over-sampling and under-sampling. Some of the top models also try different methods of data augmentation so that they can have more positive examples to tackle the imbalance issue.

By inspecting more details of submitted predictions of top systems, we found most of the wrongly classified samples require systems to understand the statement better while the existing models often lean toward memorizing and overweighting token level features to make predictions. Take a counterfactual sentence as an example, \textit{``if I were asked to, I would be happy to talk to anyone''}. This is misclassified likely because of including \textit{``were...to''} in the antecedent, which is highly correlated to non-counterfactual statements, suggesting a major flaw of existing methods. Similarly, some non-counterfactual sentences are incorrectly labelled for they include some token-level counterfactual features while not indicating counterfactuals. For example, the sentence \textit{``under the current alignment, he said, American multinational corporations like Pfizer might invest more money in the United States, not less, if they had their tax domiciles abroad''} is a non-counterfactual sentence for it is not assuming anything counter to the facts, while \textit{``if had''} part along with the modal verb (\textit{``might''} in this case) in the same sentence is usually correlated to a counterfactual.

\begin{table*}[t]
\centering
\begin{tabular}{|C{3.9cm}|C{1.8cm}|C{2cm}|C{2cm}|C{2cm}|}
\hline
\textbf{Team}     & \textbf{Ranking}
& \textbf{F1}   & \textbf{Recall}  & \textbf{Precision} \\ \hline\hline
\textit{Baseline} & - & 19.7 & 11.5 & 69.7 \\ \hline
haodingkui & 1 & 90.9 & 91.9 & 90.0 \\
josefjon   & 2 & 90.3 & 90.7 & 90.0 \\ 
Roger  & 3 & 90.0  & 88.6  & 91.5  \\ 
shngt  & 4 & 89.3  & 90.9  & 87.8   \\
lenyabloko  & 5 & 87.8  & 89.0   & 86.6   \\ 
baiyang2581 & 6 & 86.3  & 83.7 & 88.9  \\
LucasHub & 7 & 85.6  & 84.6  & 86.7   \\
Ferryman & 8 & 85.6  & 85.8  & 85.4  \\ 
pouria\_babvey & 9 & 85.6  & 85.9  & 85.3   \\
will\_go & 10 & 85.1 & 89.0  & 81.5  \\
ad6398   & 11 & 85.0 & 89.2  & 81.2  \\ 
habi-akl & 12 & 85.0 & 85.9  & 84.2  \\
rajaswa\_patil & 13  & 84.5  & 85.6  & 83.5  \\
lijunyi  & 14 & 82.9 & 82.1  & 83.7 \\ 
lidejian & 15 & 82.7 & 85.2  & 80.3  \\
changshivek & 16 & 80.2 & 90.7 & 71.9  \\
wqmike123 & 17 & 76.3 & 68.4 & 86.3 \\ 
SudeshnaJanaTCS & 18 & 72.8  & 95.7  & 58.8  \\
Serena  & 19 & 70.4 & 66.1 & 75.2  \\
Xlxw.xu & 20 & 67.6 & 89.3 & 54.4  \\ 
eldams  & 21 & 66.3 & 67.2 & 65.4  \\
yidu    & 22 & 53.8  & 38.5  & 89.3  \\
TCSNLP  & 23 & 51.9  & 97.7  & 35.3  \\ 
skblaz  & 24 & 36.5 & 53.0 & 27.8   \\
jacqle  & 25 & 33.6 & 30.6 & 37.2  \\
niji123 & 26 & 19.1 & 10.0 & 10.5  \\ 
rupsajina  & 27 & 7.2 & 5.2 & 11.7  \\ \hline

\end{tabular}
\caption{Performance of the baseline and official submissions on Subtask-1.}
\label{table: results task-1}
\end{table*}

\subsection{Subtask-2: Detecting Antecedent and Consequent (DAC)}


\label{sec: submission-2}

We build a conditional random field (CRF) model for sequence labeling as the baseline model for Subtask-2. 
This model can assign labels to each token in the input sequence by taking advantage of all input tokens and previous predictions. Specifically, same as in many name entity recognition systems, this baseline model annotates the antecedent and consequent using the B/I/O scheme, marking whether a word is at the Beginning, Inside or Outside either the antecedent or consequent. A common set of features for each word are extracted and used to train this model, including POS tags, features of nearby words, and whether the word has an uppercase/lowercase/title flag. The performance of the CRF baseline can be found in Table~\ref{table: results task-2}.


The performance of the baseline model and submitted systems are shown in Table~\ref{table: results task-2}. 
We received 11 official submissions to Subtask-2, and most of the submissions outperform the provided baseline model.
In subtask 2, top systems take two main approaches: sequence labelling or question answering, and nearly all of them benefit from pretraining. 
An exception is the 6th ranked team, \textit{Anderson\_Sung}, that use a multi-stack, birdirectional LSTM architecture to some success ~\cite{AndersonSung2020SemevalTask5}, showing that non-transformer approaches are viable for the task. Similarly, \textit{habi-akl} experiments with a BiLSTM Conditional Random Fields (CRF) model for Subtask-2, but find that a BERT based model with a multi-layer perceptron classifier outperforms the LSTM and conclude that the semi-supervised systems show a better level of understanding of challenging counterfactual forms ~\cite{habiakl2020SemevalTask5}.

One approach among the top models formulates the problem as an extractive question answering (QA) task, with the target being extracting the answer from the given context towards a specific question. The others formulate the task as a sequence labeling task. In the top 4 systems, half of the teams took the QA approach, and the other half took the sequence labelling approach. The choice between BERT and RoBERTa has split almost evenly amongst most of the participants. As is the case for Subtask-1, many teams sought to build on top of the pre-trained models and add additional upper layer structures to handle the task better.

The best results are achieved by team \textit{Martin}, with an F1 score of 88.2 and an exact match score of 57.5 ~\cite{josefjon2020SemevalTask5}. To predict the start and ending positions of antecedents and consequents, the model utilizes an ensemble of RoBERTa models and extend it in the same manner as how BERT was extended for the SQuAD dataset~\cite{rajpurkar2016squad}. 
The second-place system \textit{pouria\_babvey} uses a sequence labelling approach: the authors develop the model on top of BERT with a multi-head attention layer and label masking to capture mutual information between nearby labels ~\cite{pouriababvey2020SemevalTask5}. Label masking, in which only part of the labels is fed during training and the rest have to be predicted, has shown to be particularly effective for improving accuracy, which can be seen as a form of regularization. In addition, a multi-stage algorithm is used to gradually improve certainty in predictions after each step. 
The third-place system, ~\textit{Roger}, formulates the problem as a query-based question answering problem, where antecedents and consequents are extracted after an antecedent and consequent query are supplied along with the original statement into BERT. Pointer networks are further used to predict the start and ending positions ~\cite{Roger2020SemevalTask5} . 
A unique approach for Subtask-2 is used by 7th placed team, \textit{rajaswa\_patil}, where they use a base architecture for both subtasks. They first train with a binary-classification module for Subtask-1, then replace it with a regression-module and further fine-tune the system for Subtask-2 ~\cite{rajaswapatil2020SemevalTask5}, leveraging the commonality between the two tasks.

We can observe that there is still a gap between the performance of exact match and F1, which is mainly due to the fact that Exact Match is sensitive to non-essential phrases in predictions even the core parts are identified correctly.

\begin{table*}[t]
\centering
\begin{tabular}{|C{3.3cm}|C{1.8cm}|C{2cm}|C{2cm}|C{2cm}|C{2cm}|}
\hline
\textbf{Team} &\textbf{Ranking} & \textbf{F1} & \textbf{Recall} & \textbf{Precision} & \textbf{Exact Match} \\ \hline\hline
\textit{Baseline} & - &55.5 &54.9 &56.8 &34.3 \\ \hline
Martin & 1 & 88.2 & 89.3 & 90.0 & 57.5  \\
pouria\_babvey & 2 &87.8 & 87.5 & 91.3  & 49.7 \\ 
Roger & 3 & 87.5  & 90.8  & 87.5  & 54.6  \\ 
ywzhang & 4 & 84.1 & 84.6 & 86.8  & 47.1  \\
habi-akl & 5 & 83.9 & 88.8 & 82.3 & 25.9  \\ 
Anderson\_Sung  & 6 & 78.4 & 81.2 & 80.9 & 28.2 \\
rajaswa\_patil & 7 & 68.8 & 67.2 & 74.0   & 0.0  \\
lidejian & 8 & 60.6  & 63.3 & 60.7 & 0.1 \\ 
aniojha  & 9 &48.3 & 51.8  & 47.1 & 3.12 \\
Xlxw.xu & 10 & 47.6  & 50.9 & 47.2  & 0.0 \\
ElvisInalco  & 11 & 8.9 & 26.3 & 6.0  & 0.4 \\ 

\hline

\end{tabular}
\caption{
Performance of the baseline and official submissions on Subtask-2.}
\label{table: results task-2}
\end{table*}

\section{Related Work}
\label{sec: related work}
Modelling counterfactual thinking has started to attract more interest. One of the previous works closest to ours is~\cite{son2017recognizing}, in which a small-scale counterfactual tweet dataset is collected from social media. There are three main differences between that dataset and ours. First, there are only 2,000 samples in the tweet dataset (including the supplement data mentioned in the paper), while our dataset for counterfactual detection in Subtask-1 is ten times larger, which we believe is important for training deep learning based models.
Second, our benchmark provides evaluation for antecedents and consequents extraction, which are essential components of counterfactual analysis. Third, our dataset includes statements from three different domains (finance, politics, healthcare). In contrast to the statements collected from tweets, which have a very large portion that are open-ended, vague thoughts, the counterfactuals in our dataset are more meaningful domain-related statements.

There is another dataset~\textit{TIMETRAVEL} proposed in~\cite{qin2019counterfactual} for counterfactual story rewriting, in which given a short story and an alternative counterfactual event context, the story needs to be minimally revised to keep compatible with the intervening counterfactual event. The empirical results show that it is still challenging for current neural language models to perform well on the counterfactual story rewriting task due to the lack of counterfactual reasoning capabilities.

In a broader viewpoint, counterfatuals are an important form of causal reasoning. Researchers argue that the notion of counterfactuals is essential for causal reasoning, in which causal modeling is proposed to interpret counterfactual conditionals in natural language, and such work has been discussed since the possible worlds semantics developed in the 1970s~\cite{lewis2013counterfactuals,lewis1986causation}. 
The more recent work renders useful insights by formulating causal inference as a three-level hierarchy, which are \textit{association}, \textit{intervention}, and \textit{counterfactual}, respectively~\cite{pearl2018book,pearl2019seven}. The top of the hierarchy is counterfactual---if a model can correctly answer counterfactual queries like \textit{``what would happen if we had acted differently''}, it should also be able to answer association and intervention queries.
The research in ~\cite{pearl1995causal,pearl2010introduction} also made contributions to a general theory of causal inference, which is based on the Structural Causal Model (SCM), and counterfactual analysis is provided with a formal mathematical formalism. 



\section{Summary and Future Work}
\label{sec: conclusion}
We present a counterfactual recognition task that includes two basic subtasks. Subtask-1 evaluates whether a given statement is counterfactual or not with 20,000 training and test statements. Subtask-2 aims at recognizing antecedents and consequents in counterfactual statements. The official task received 27 submissions to Subtask-1 and 11 submissions for Subtask-2. The state-of-the-art performances achieved a 90\% F1 score in Subtask 1, as well as an 88.2\% F1 and 57.5\% Exact Match score in Subtask-2. We hope this task and dataset will intrigue and facilitate further research on counterfactual analysis in natural language. 

\section*{Acknowledgements}
The first author was supported by Vector Institute Scholarship on Artificial Intelligence and Alibaba Innovative Research Grant. The second author was supported NSERC Undergraduate Student Research Awards (USRA). The last author is sponsored by NSERC Discovery Grants and DAS Grants. We thank Jiaqi Li and Qianyu Zhang for their help in this project.

\bibliographystyle{coling}
\bibliography{semeval2020}

\newpage
\section*{Appendix A. Token-based Patterns}
\label{token-based-patterns}
The full list of token-based patterns are listed in Table~\ref{table: templates}. 
Specifically we have 14 rows of patterns. On each raw, we first apply the inclusion patterns listed in the second column to identify candidate counterfactual statements, and then we apply the additional exclusion patterns in the third column to remove the candidates that satisfy these exclusion rules.







\section*{Appendix B. POS-based Filtering Rules} 
The POS-based filtering patterns are only applied to candidate counterfactuals found by the rules on row 1, 8, 11, and 14 in Table~\ref{table: templates}. Specifically, for candidates found by these rules, only if they further satisfy the POS-based patterns, they will be finally included. When we applying the following rules, words that match a set of POS tag criteria must be present somewhere in a sentence. Although the matched words do not have to directly follow one another and other words can be in between, they have to be in a particular order. The POS-based rules and their order are detailed as follows:

\begin{table*}[htbp]
\centering
\begin{tabular}{|p{0.8cm}|p{7cm}|p{7cm}|}
\hline
Index &  Inclusion Patterns      & Additional Exclusion Patterns    \\ \hline
1 & if ... then ... & even/what/as if ...  then ...     \\ \hline
2 & if ... had/hadn't/had not ... 
& even/what/as if ... had/hadn't/had not ... \newline OR any sentences where a/the/to/an immediately follows had/hadn't/had not   \\\hline
3 & 'd/could/may/might/should/would/ought to have/haven't/not have ... \newline OR wouldn't/couldn't/shouldn't have ...  & Sentences in which a/an/the/to immediately follows have/not have/haven't  \\\hline
4 & what if ... & N/A \\\hline
5 & even if ... & N/A \\\hline
6 & if I/there/he/she/you were/weren't/were not ...  & even/what/as if I/there/you/he/she/you were/weren't/were not ... \newline OR Sentences where ``to'' follows were/weren't/were not \\ \hline
6 & if ... were/weren't/were not to ...  & even/what/as if ... were/weren't/were not to ... \\\hline
7 & wish ... could/may/should/wouldn't/couldn't/  \newline shouldn't have/not have/haven't ...  \newline OR wish I'd/we'd/you'd/he'd/she'd/they'd/\newline there'd have/not have/haven't ...  & Sentences that fit either of the patterns but have ``to'' immediately follow ``wish'' or a/the/to/an immediately follow have/haven't  \\\hline
8 & wish ... were/weren't/had/had not/hadn't ...  &  Sentences that fit the pattern but have ``to'' immediately follow ``wish'' or a/the/to/an follow were/weren't/had/had not/hadn't \\\hline
9 & wish ... & Sentences where ``to'' immediately follows ``wish'' \\\hline
10 & but for ... could/might/would/should/wouldn't/  \newline couldn't/shouldn't have/not have/haven't ... & Sentences that follow the pattern but where ``now'' follows ``but for'' and a/the/to/an  follows have/haven't/have not   \\\hline
11 & if only ... & even/what/as if only ... \newline OR if only for ... \\\hline
12 & had/were ... & had/were ... ? \\\hline
13 & if ... & even/what/as if \\\hline
14 & I'd/we'd/you'd/he'd/she'd/they'd/there'd/\newline would/could/should/might/wouldn't/couldn't/ \newline shouldn't have ... without ... \newline OR without ... I'd/we'd/you'd/he'd/\newline she'd/they'd/there'd/would/could/should/\newline might/wouldn't/couldn't/shouldn't/would not/could not/should not have ...  & Sentence cannot end with a question mark or exclamation mark \\
\hline
\end{tabular}
\caption{
Token patterns used in the filtering stage. On each raw, we first apply the patterns listed in the second column to identify candidate counterfactual statements, and then we apply the patterns in the third column to additionally remove the candidates just found.
}
\label{table: templates}
\end{table*}

\begin{enumerate}
  \item \textit{Conjunctive normal}: This is a counterfactual form in which the consequent follows the antecedent. The~\textit{conjunctive normal} form dictates a conditional conjunction is followed by a past tense subjective or past modal verb in the antecedent, and there is a past or present tense modal verb in the consequent~\cite{janockocounterfactuals}. To filter sentences following this form, we first search for the word \textit{if} in the sentence. A word with a past tense verb, modal verb, or past participle verb speech tag in the sentence must then follow. Lastly, a word with a modal verb tag must come after. 
  
  \item  \textit{Conjunctive Converse}: 
  An antecedent follows the consequent in a \textit{conjunctive converse} form sentence.
  In the consequent, there has to be a modal verb followed by a past or present tense verb. In the antecedent, there has to be a conditional conjunction followed by a past tense subjective or past tense modal verb~\cite{janockocounterfactuals}.
  Locating counterfactual sentences of this form requires locating sentences that contain a token with a modal verb tag, and are followed by a token with a base form verb tag. An \textit{if} has to be present afterwards, and after this a token needs to contain either a modal verb, past tense verb, or past participle verb token. 
  
  \item  \textit{Modal Normal}: A modal normal counterfactual sentence has the consequent following the antecedent. Inside the antecedent, there must be a modal verb and a past participle verb, and inside the consequent there must be past/present tense modal verb~\cite{janockocounterfactuals}. Capturing this form entails selecting sentences that have a token with a modal verb tag. A token containing a past participle verb tag must be somewhere after, and then there must be a token with a modal verb tag.
  
  \item \textit{Wish/Should Implied}: The \textit{Wish/Should Implied counterfactual} form only explicitly contains an antecedent in the sentence, with the consequent being implied, and it must contain an independent clause following a wish or should~\cite{janockocounterfactuals}. To capture this form, sentences that contain the token 'wish' and that have a word after this with a past tense verb or a past participle verb tag.
  \item \textit{Verb Inversion}: The category has two specific forms that differ in if the antecedent presents before or after the consequent. In either case, according to ~\cite{janockocounterfactuals}, the antecedent contains a \textit{had} or \textit{were} inversion along with a past tense verb, and the consequent has a modal verb and a past or present tense verb.  
    \begin{enumerate}
    \item The antecedent presents first in this case. Thus, for this form the sentence first has to contain \textit{had} or \textit{were} as the first token. After this, a token with a past tense or past participle verb token must be present, but only if the first token is \textit{had}. In either case, a word with a modal verb tag has to follow, and then further followed by a base verb, present non third person singular verb, present third person singular verb, or a past tense verb tag.
    \item In this form a consequent presents first. As a result, a word with a modal verb tag must follow, and is in turn followed by a word with base verb, present non third person singular verb, present third person singular verb, or a past tense verb tag. After this the sentence had to contain a \textit{had} or \textit{were} token. If it does contain a \textit{had} token, an additional past tense or past participle verb token word has to also follow it.
    \end{enumerate}
\end{enumerate}

\section*{Appendix C. Word Count Frequency Statistics}
\begin{figure}[!h]
  \centering
  \includegraphics
[width=\linewidth,trim={0.1cm 0.1cm 0.1cm 0.1cm},clip]{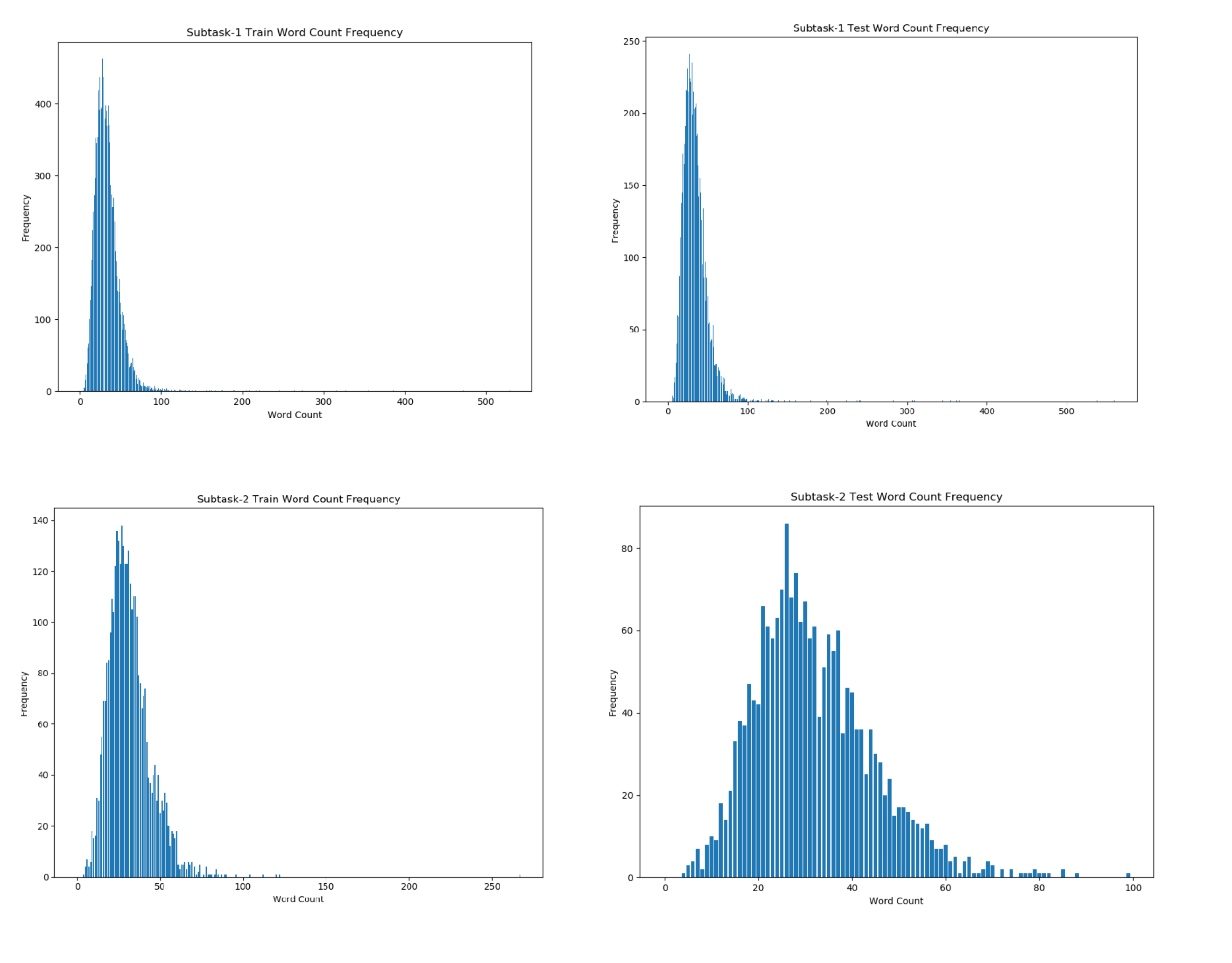}
  \caption{Statistics of samples in Subtask-1 and Subtask-2.}
\label{fig:statistics}
\end{figure}

\end{document}